\documentclass[10pt,twocolumn,letterpaper]{article}

\usepackage{cvpr}
\usepackage{times}
\usepackage{epsfig}
\usepackage{graphicx}
\usepackage{amsmath}
\usepackage{amssymb}
\usepackage{booktabs}
\usepackage{multirow}
\usepackage{authblk}
\usepackage[activate={true,nocompatibility},final,tracking=true,kerning=true,spacing=true,factor=1100,stretch=10,shrink=10]{microtype}

\DeclareMathOperator*{\argmin}{arg\,min}

\usepackage[pagebackref=true,breaklinks=true,letterpaper=true,colorlinks,bookmarks=false]{hyperref}

\cvprfinalcopy 

\def\cvprPaperID{1455} 

\newcommand{\squeezeupSmall}{\vspace{-2mm}}
\newcommand{\squeezeup}{\vspace{-4mm}}

\ifcvprfinal\fi

\begin{document}

\title{LiveSketch: Query Perturbations for Guided Sketch-based Visual Search}


\author[1 2]{John Collomosse}
\author[1]{Tu Bui}
\author[2]{Hailin Jin}
\affil[1]{Centre for Vision Speech and Signal Processing, University of Surrey}
\affil[2]{Creative Intelligence Lab, Adobe Research }


\maketitle
\thispagestyle{empty}

\begin{abstract}
LiveSketch is a novel algorithm for searching large image collections using hand-sketched queries.  LiveSketch tackles the inherent ambiguity of sketch search by creating visual suggestions that augment the query as it is drawn, making query specification an iterative rather than one-shot process that helps disambiguate users' search intent. Our technical contributions are: a triplet convnet architecture that incorporates an RNN based variational autoencoder to search for images using vector (stroke-based) queries; real-time clustering to identify likely search intents (and so, targets within the search embedding); and the use of backpropagation from those targets to perturb the input stroke sequence, so suggesting alterations to the query in order to guide the search.  We show improvements in accuracy and time-to-task over contemporary baselines using a 67M image corpus.
\end{abstract}

\section{Introduction}

Determining user intent from a visual search query remains an open challenge, particularly in sketch based image retrieval (SBIR) over millions of images where a sketched shape can yield plausible yet unexpected matches. For example, a user's sketch of a dog might return a map of the United States that ostensibly resembles the shape ({\em structure}) drawn, but is not relevant. Free-hand sketches are often incomplete and ambiguous descriptions of desired image content \cite{Collomosse2008}.  This limits the ability of sketch to communicate search intent, particularly over large image datasets.

\begin{figure}[t!]
    \centering
    \includegraphics[width=1.0\linewidth]{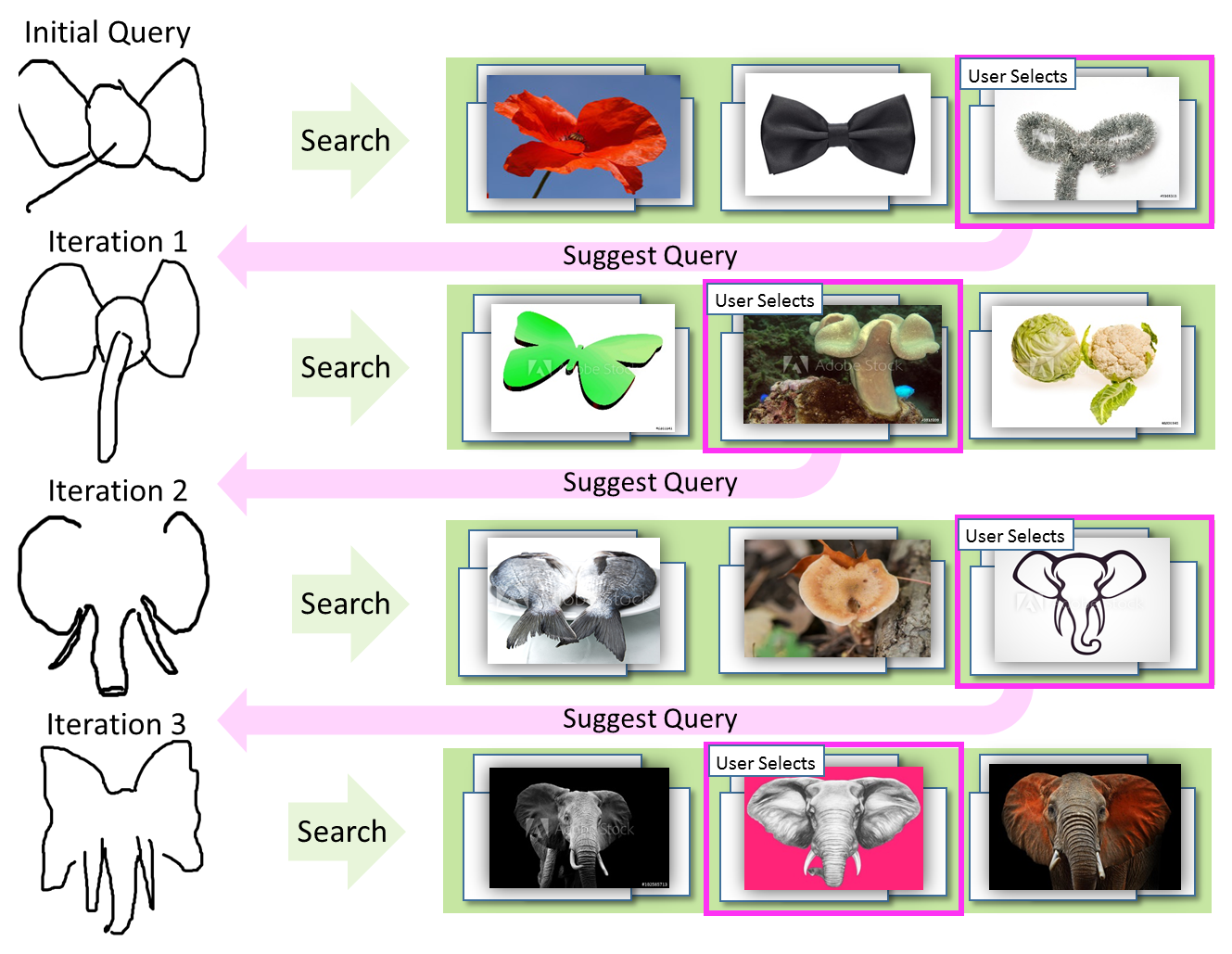}
    \caption{LiveSketch helps disambiguate SBIR for large datasets, where a shape sketched from scratch (top left) can yield results that do not match the users' search intent. LiveSketch iteratively suggests refinements to users' sketched queries to guide the search (Iter.1-3), based on the user indicating relevant clusters of results (right).  This interaction disambiguates and quickly guides the search towards results that match users' search intent  (subsec.~\ref{sec:eval_system}).
    }\vspace{-2em} \label{fig:teaser}
\end{figure}


This paper proposes LiveSketch; a novel interactive SBIR technique in which  users iterate to refine their sketched query, selecting and integrating sketch embellishments suggested by the system in order to {\em disambiguate search intent} and so improve the relevance of results (Fig.~\ref{fig:teaser}). A core  novelty of our approach lies within the method by which visual suggestions are generated, exploiting the reversibility of deep neural networks (DNNs) that are commonly used to encode image features to create the search index in visual search systems \cite{CTU-ECCV2016,gordo2016deep,sketchstyle2017,Bui2018}.  By identifying clusters of likely target intents for the user's search, we reverse the DNN encoder to ‘explain’ how such clusters could be generated by adapting the query.  We are inspired by adversarial perturbations (APs); that use backpropagation to generate `adversarial' image examples \cite{Goodfellow2015} that induce object mis-classification \cite{Dezfooli2017,Athalye2017,WhyNotWorry} to a targeted category.  In our context of visual search, we similarly backpropagate to perturb the sketched query from its current state toward one (or more) targets identified in the search embedding by the user.  As such, the query becomes a `living sketch' on the canvas that reacts interactively to intents expressed  by the user, forming the basis for subsequent search iterations.  The use of a single, live sketch to collaboratively guide the search differs from prior approaches such as ShadowDraw \cite{shadowdraw} that ghost hundreds of top results on the canvas.  We propose three  technical contributions:\\
\noindent{\bf1) Vector Queries for Sketch based Image Retrieval.} We learn a joint search embedding that unifies vector graphic and raster representations of visual structure, encoded by recurrent (RNN) and convnet (CNN) branches of a novel triplet DNN architecture.  
Uniquely, this embedding enables the retrieval of raster (\eg photo) content using sketched queries encoded  as a sequence of strokes.  This higher level representation is shown to not only enhance search accuracy (subsec.~\ref{sec:eval_embed}) but also enables perturbation of the query to form  suggestions, without need for pixel regularization.\\
\noindent{\bf2) Guided Discovery of Search Intent.} We make use of an auxiliary (semantic) embedding to cluster search results into pools, each representing a candidate search intent. For example, a circle on a stick might return clusters corresponding to balloons, signs, mushrooms. Deriving query suggestions from sketches drawn from these pools guides the user toward relevant content, clarifying intent by supplying contextual information not present in the query.\\
\noindent{\bf3) Query Perturbation.} We propose an iterative strategy for SBIR query refinement in which the users' query sketch is perturbed to incorporate the appearance of search intent(s) indicated by the user. We cast this as a search for a query perturbation that shifts the encoded query within the search embedding closer toward those selected intent(s),  encoding that vector as a loss (in the spirit of APs) that is backpropagated through the DNN to update the sketch.

\section{Related Work}

Visual search is a long-standing problem within the computer vision and information retrieval communities, where the iterative presentation and refinement of results has been studied extensively as ‘relevance feedback’ (RF) \cite{Setia2005,Kovashka2012,Kovashka2013} although only sparsely for SBIR \cite{James2014b}.  RF is driven by interactive markup of results at each search iteration. Users tag results as relevant or irrelevant, so tuning internal search parameters to improve results.  Our work differs in that we modify the query itself to affect subsequent search iterations; queries may be further augmented by the user at each iteration. Recognizing the ambiguity present in sketched queries we group putative results into semantic clusters and propose edits to the search query for each object class present. 

Query expansion (QE) is a automated technique to improve search accuracy from a single, one-off visual query \cite{Philbin2007,Bhattacharjee2016,Tolias2017,Radenovic2018} by recursively submitting search results as queries. LiveSketch contrasts with QE as it is an {\em interactive} system in which query refinements are suggested, and optionally incorporated by the user to help disambiguate search intent; so communicating more information than present in a single, initial sketched query.
  
Deep learning, specifically CNNs (convnets), have been rapidly adopted for SBIR and more broadly for visual search outperforming classical dictionary learning based models (e.g. bag of words) \cite{Sivic2003,Bui2015}.  Wang et al \cite{wang2015sketch} were arguably the first to explore CNNs for sketched 3D model retrieval via a contrastive loss network mapping sketches to rendered 2D views. Qi \etal \cite{Qi2016} similarly learned correspondence between sketches and edge maps. Fine-grained SBIR was explored by Yu \etal \cite{Yu2016} and Sangkloy \etal \cite{Hays2016} who used a three-branch CNN with triplet loss for learning the cross-domain embedding.  Triplet loss models have been used more broadly for visual search e.g. using photographic queries \cite{Wang2014,CTU-ECCV2016,gordo2016deep}.  Bui \etal \cite{BuiArxiv2016,Bui2018} perform cross-category retrieval using a triplet model and currently lead the Flickr15k \cite{Hu2013}  benchmark for SBIR.   Their system was combined with a learned model of visual aesthetics \cite{bam} to constrain SBIR using stylistic cues in \cite{sketchstyle2017}.  All of these prior techniques learn a deep encoder function that maps an image into a point in a metric search embedding where the distance between an image pair correlates to its similarity.  Such embeddings can be binarized (\eg via PQ \cite{Jegou2010}) for scalable search.  In prior work search embeddings were learned using rasterized sketches i.e. images, rather than vector representations of sketched strokes.   In our approach we adopt the a vector representation for sketches, building upon the SketchRNN variational auto-encoder of Eck \etal previously applied to blend \cite{Eck2018} and match \cite{Xu2018} sketches with sketches.  Here we adapt SketchRNN  in a more general form for  both our interactive search of photographs, and for generating search suggestions, training with the Quickdraw50M dataset \cite{qd}.

Our work is aligned to ShadowDraw \cite{shadowdraw} in which ghosts (edge-maps derived from top search results) are averaged and overlaid onto the sketch canvas (similarly, \cite{Efros2014} for photo search).  However our system differs both in intent and in method.  ShadowDraw is intended to teach unskilled users to sketch rather than as a search system in its own right \cite{shadowdraw}.  The technical method also differs -– our system uses deep neural networks (DNNs) both for search and for query guidance, and hallucinates a single manipulable sketch rather than a non-edittable cloud of averaged suggestions.  This declutters presentation of the suggestions and does not constrain  suggestions to the space of existing images in the dataset.  Our method produces query suggestions by identifying destination points within the search embedding and employing backpropagation through the deep network (with network weights fixed) in order to update the input query so that it maps to those destination points.  The manipulation of input imagery with the goal of affecting change in the output embedding is common in the context of adversarial perturbations (APs) where image pixels are altered to change the classification (softmax) output of a CNN \cite{Dezfooli2017,Athalye2017}.  We are inspired by FGSM \cite{Goodfellow2015} which directly backpropagates from classification loss to input pixels in order to induce noise that, whilst near-imperceptible, causes mis-classification with high confidence.  Although we also  backpropagate, our goal differs in that we aim for observable changes to the query that  guide the user in refining their input.  Our reimagining of APs for interactive query refinement in visual search is unique.

\section{Methodology}
\label{sec:method}

\begin{figure}[t!]
    \centering
    \includegraphics[width=1.0\linewidth]{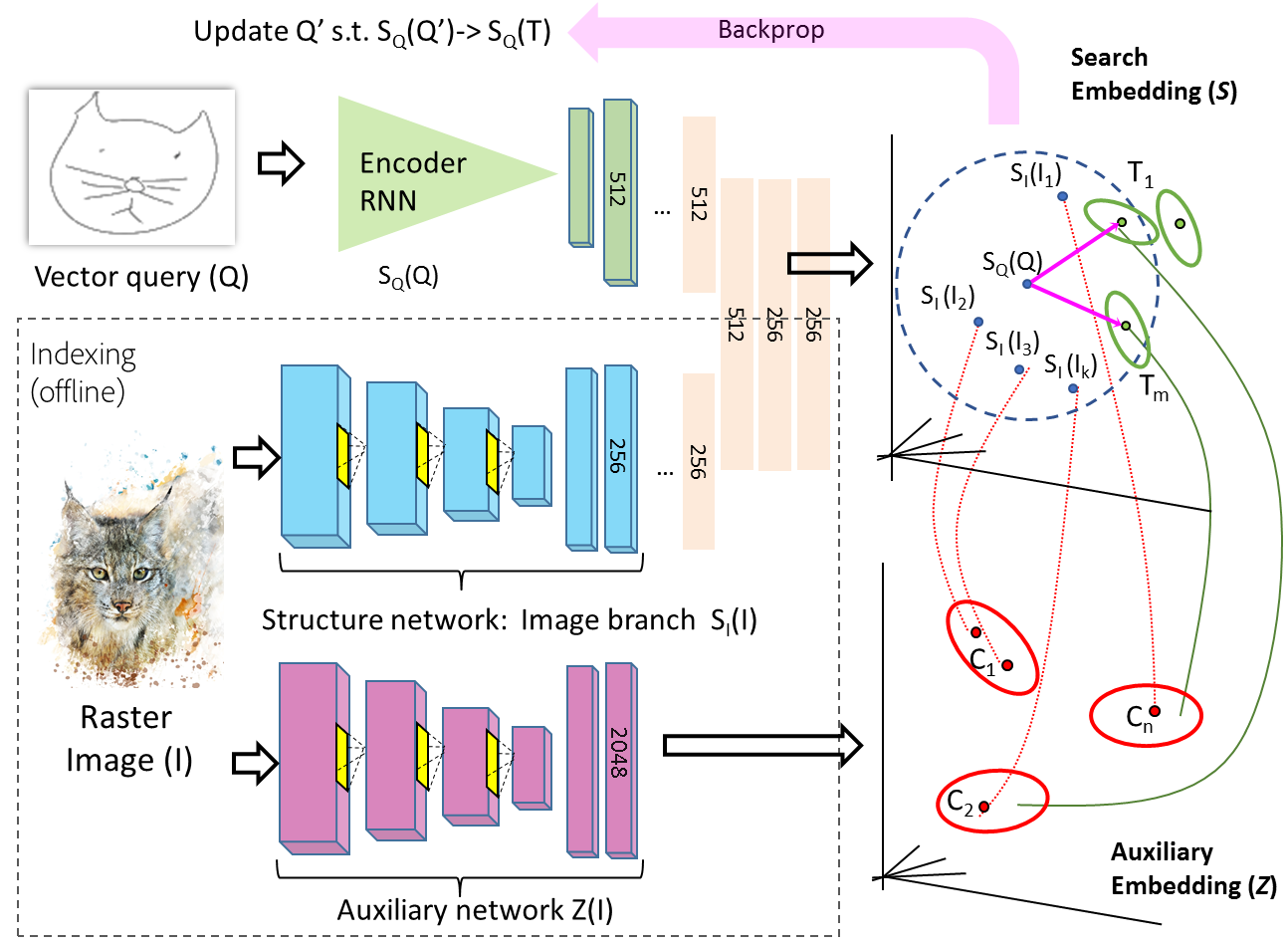}
    \caption{Overview of the proposed SBIR framework.  A query sketch ($Q$, vector graphics form) and images ($I$, raster form)  are encoded into the search embedding $\mathcal{S}$ via RNN and CNN branches, unified via four inner product layers. Images are encoded via $S_I(.)$; the image branch of \cite{Bui2018}. Query sketches are encoded via $S_Q(.)$; the encoder stage of Fig.~\ref{fig:sketchrnn}.  An auxiliary semantic embedding $\mathcal{Z}$ clusters  results to help the user pick search target(s) $T$ in the search embedding. In the spirit of adversarial perturbation, the strokes $Q$ are adjusted to minimize $||S_Q(Q)-T_i||_2$ and so evolve the sketch toward the selected target(s).}
 \label{fig:overview}
 \squeezeupSmall
\end{figure}

LiveSketch accepts a query sketch $Q$  in vector graphics form (as a variable length sequence of strokes), and searches a large ($\sim 10^8$) dataset of raster images $I=\{I_1,...,I_N\}$.  Our two-stream network architecture (Fig.~\ref{fig:overview}) unifies both vector and raster modalities via a common search embedding ($\mathcal{S}$).  Sketch and image content are encoded via RNN and  CNN branches respectively, unified via 4 fully connected (fc) layers;  final layer activations yield $\mathcal{S} \in \Re^{256}$. The end-to-end RNN and CNN paths through the network describe the pair of encoding functions $S_Q(Q)$ and $S_I(I_i)$ for encoding the visual structure of sketches, and of images, respectively; the process for learning these functions is described in subsec.~\ref{sec:training}.  Once learned, the image dataset is indexed (offline) by feeding forward all $I_i \in I$ through $S_I(.)$.  At query time, results for a given $Q$ are obtained by ranking on  $||S_Q(Q) - S_I(I_i)||_2$  where $||.||_2$ is the $L_2$ norm.

Fig.~\ref{fig:overview} provides an overview of our interactive search.  Given an initial query $Q$, images embedded in $\mathcal{S}$ proximate to $S_Q(Q)$ are returned. Whilst these images  share the visual structure of $Q$, the inherent ambiguity of sketch typically results in semantically diverse content, only a subset of which is relevant to the user's search intent.  We therefore invite the user to disambiguate their sketch intent via interaction.  Search results are clustered within an `auxiliary' semantic embedding $\mathcal{Z}$.  The user assigns relevance weights to a few ($m=3$) dominant clusters.  For each  cluster $\{C_1,...,C_m\}$ in $\mathcal{Z}$, a search target $\{T_1,...,T_m\}$ is identified in $\mathcal{S}$ (process described in subsec.~\ref{sec:intent}).  The targets receiving high weighting from the user represent visual structures that we will evolve the existing query sketch $Q$ toward, in order to form a query suggestion ($Q'$) to guide the  next search iteration.

Our query is represented in vector graphics form to enable suggestions to be generated via direct modification of the stroke sequence encoded by $Q$, avoiding the need for complex pixel-domain regularization.  LiveSketch updates $Q \mapsto Q'$ such that $S_Q(Q')$ is closer to targets $\{T_1,...,T_m\}$ than $S_Q(Q)$. Treating the weighted distances between those targets and $S_Q(Q')$ as a loss, we fix $S_Q(.)$ and propagate gradients back via the RNN branch to perturb the input sequence of strokes (subsec.~\ref{sec:perturb}) and so suggest the modified sketch query.  $Q'$ may be further augmented by the user, and  submitted for a further iteration of search.

\subsection{Cross-modal Search Embedding ($\mathcal{S}$)}
\label{sec:training}

We wish to learn a cross-modal search embedding in which a sketched query expressed as a variable length sequence of strokes, and an image indexed by the system  (\eg a photograph) containing similar visual structure, map to similar points within that embedding.  We learn this representation using a triplet network (Fig.~\ref{fig:training}) comprising an RNN anchor (a) and siamese (\ie identical, shared weights) positive and negative CNN branches (p/n). The RNN and CNN branches encode vector and raster content to intermediate embeddings $\mathcal{V}$ and $\mathcal{R}$ respectively; we describe how these are learned in subsecs~\ref{sec:sketchrnn}-\ref{sec:strucnet}. The branches are unified by 4 fully-connected (fc) layers, with weight-sharing across all but the first layer to yield the common search embedding $\mathcal{S}$.  Thus the fc layers encode two functions mapping $\mathcal{V} \mapsto \mathcal{S}$ and $\mathcal{R} \mapsto \mathcal{S}$ respectively; we write these $F_V(.)$ and $F_R(.)$ and in subsec.~\ref{sec:learnfc} describe incorporation of these into the pair of end-to-end encoding functions $S_Q(.)$ and $S_I(.)$ for our network (Fig.~\ref{fig:overview}).

\begin{table}[t!]
{
\centering
~~~~~~~~~~~~\begin{tabular}{|l|l|}
\toprule 
$\mathcal{R} \in \Re^{256}$ & Raster embedding \cite{Bui2018}*\\
$\mathcal{V} \in \Re^{512}$ & Vector graphics embedding*\\
$\mathcal{S} \in \Re^{256}$ & Joint search  embedding (structure)\\
$\mathcal{Z} \in \Re^{2048}$ & Auxiliary embedding (semantic)\\
\bottomrule
\end{tabular}
}
\caption{Summary of the feature embeddings used in LiveSketch; * indicates intermediate embeddings not used in the search index.}
\vspace{-1em}
\label{tbl:glossary}
\end{table}

\begin{figure}[b!]
    \centering
    \includegraphics[width=1.0\linewidth]{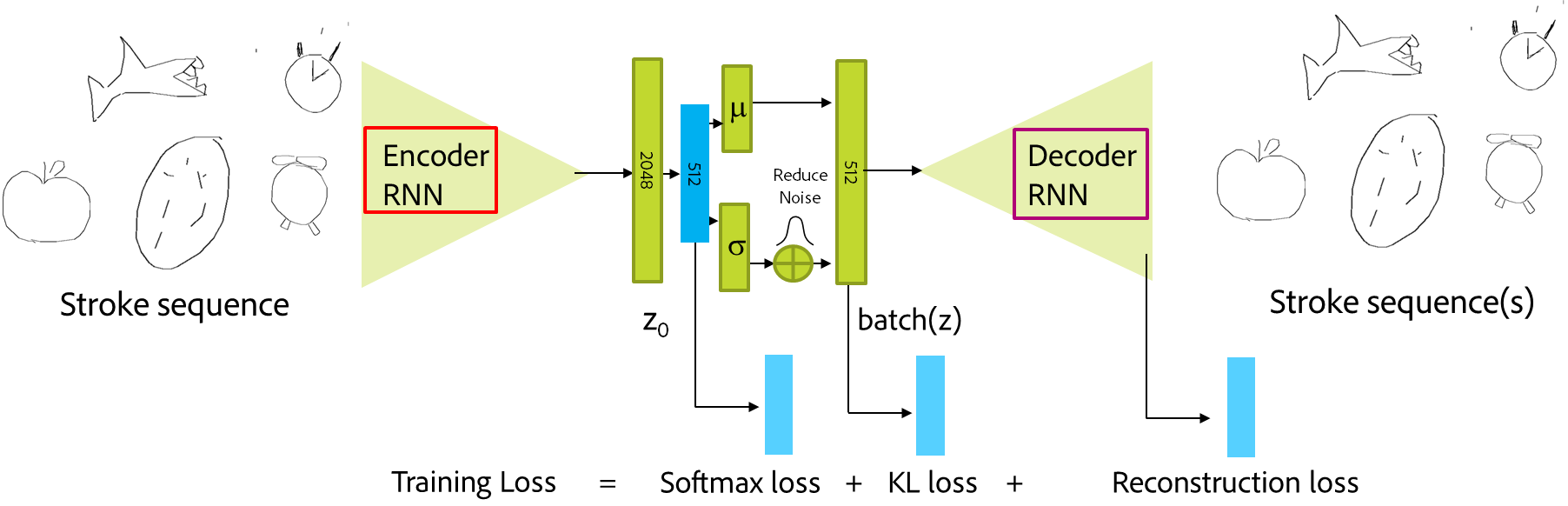}
    \caption{Modified SketchRNN \cite{Eck2018} (changes, blue) used to encode/decode stroke sequences via addition of 512-D latent representation and classification loss. Integrates with Fig.~\ref{fig:overview} (anchor).}
    \label{fig:sketchrnn}
\end{figure}
\subsubsection{Sketch Variational Autoencoder}
\label{sec:sketchrnn}
The RNN branch is a forward-backward LSTM encoder comprising the front half of a variational autoencoder (v.a.e.) for sketch encoding-decoding, adapted from the SketchRNN network of Eck \etal \cite{Eck2018}.  In the SketchRNN v.a.e., a deterministic latent representation ($z_0$) is learned, alongside parameters of multi-variate Gaussian from which a non-deterministic (n.d.) representation ($\mathrm{batch_z}$) is sampled to drive the decoder and reconstruct the sketch through recurrence conditioned on $\mathrm{batch_z}$.  The representation is learned via a combination of reconstruction loss (and a regularization `KL loss'~\cite{EckKL2017} over the multi-variate parameters), but as proposed \cite{Eck2018} can represent only up to  a few object classes making it unsuitable for web-scale SBIR.  

We adapt SketchRNN as follows (Fig.~\ref{fig:sketchrnn}). We retrain from scratch using 3.5M sketches from Quickdraw50M (QD-3.5M; see Sec.~\ref{sec:results}) adding a low-dimensional (512-D) bottleneck  after $z_0$ from which $\mathrm{batch_z}$ is sampled.  We add softmax classification loss to that bottleneck, weighted equally with the original reconstruction and KL loss terms. During training we  reduce below $10^{-2}$ the covariance of the n.d. variate.  A query sketch ($Q$) is coded as a sequence of 3-tuples $Q=\left[q_1, q_2, ... , q_n\right]$ where $q_i=\left(\delta x, \delta y, l\right)$ representing relative pen movements in ${x,y} \in \Re^2$ and whether the pen is lifted $l=[0,1]$; an abbreviated form of the 5-tuple coding  in \cite{Eck2018}.  The intermediate embedding available at the bottleneck ($\mathcal{V} \in \Re^{512}$) is capable of reconstructing sketches across diverse object classes (c.f. Sec.\ref{sec:eval_interp}).  The encoder forms the anchor of the  proposed triplet network (Fig.~\ref{fig:training}); we denote the encoding and decoding functions as $V_{E}(Q) \mapsto \mathcal{V}$ and $V_{D}(\mathcal{V}) \mapsto Q$.  

\subsubsection{Raster Structure Encoder}
\label{sec:strucnet}

To encode raster content, we adopt the architecture of Bui \etal \cite{Bui2018} for the CNN branch.  Their work employs a triplet network with GoogLeNet Inception backbone \cite{Googlenet2015} that unifies sketches (in raster form) and images within a joint search embedding. One important property is the partial sharing of weights between the sketch CNN branch (anchor) and the siamese image CNN (+/-) branches of their triplet network.  Once trained, these branches yield two functions: $R_S(.)$ and $R_I(.)$, that map sketched and image content  to a joint search embedding.  Full details on the multi-stage training of this model are available in \cite{Bui2018}; we use their pre-trained model in our work and incorporate their joint embedding as the intermediate embedding $\mathcal{R} \in \Re^{256}$ in our work.  Specifically, $R_S(.)$ is used to train our model (Fig.~\ref{fig:training}, p/n).

\subsubsection{Training the Joint Search Embedding}
\label{sec:learnfc}
\begin{figure}[t!]
    \centering
    \includegraphics[width=1.0\linewidth]{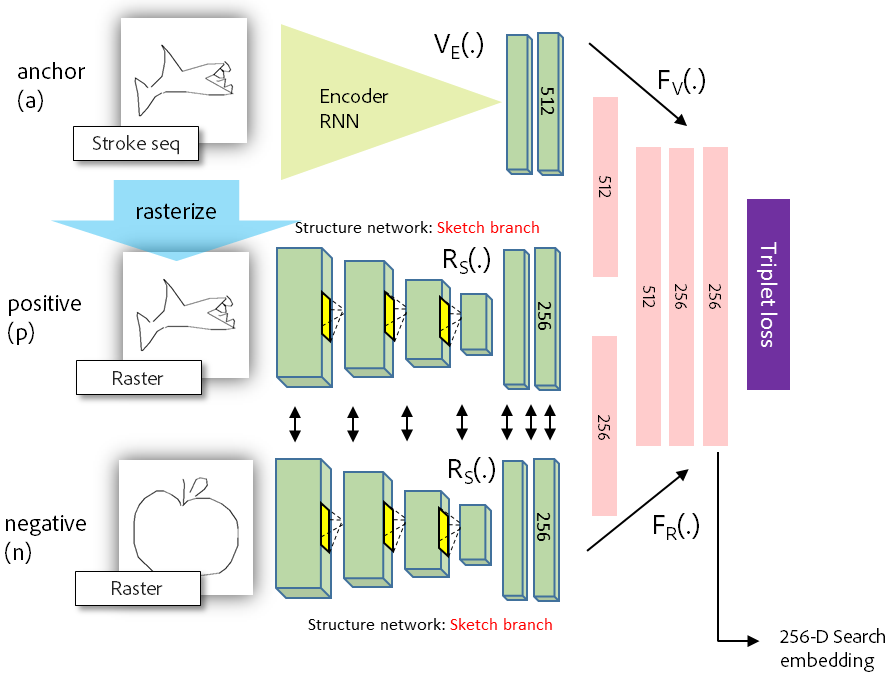}
    \caption{Training the LiveSketch network; an encoder that maps raster and vector content to a common search embedding.   The search embedding is trained using raster and vector (stroke sequence) content sampled from QD-3.5M.  During training, the CNN branches (p/n) are $R_S(.)$ \ie the sketch branch of \cite{Bui2018}. However branch $R_I(.)$ is used at inference time (Fig.~\ref{fig:overview}).}
    \label{fig:training}
    \squeezeup
\end{figure}

\begin{figure*}[t!]
    \centering
    (a)
    \includegraphics[height=3.75cm]{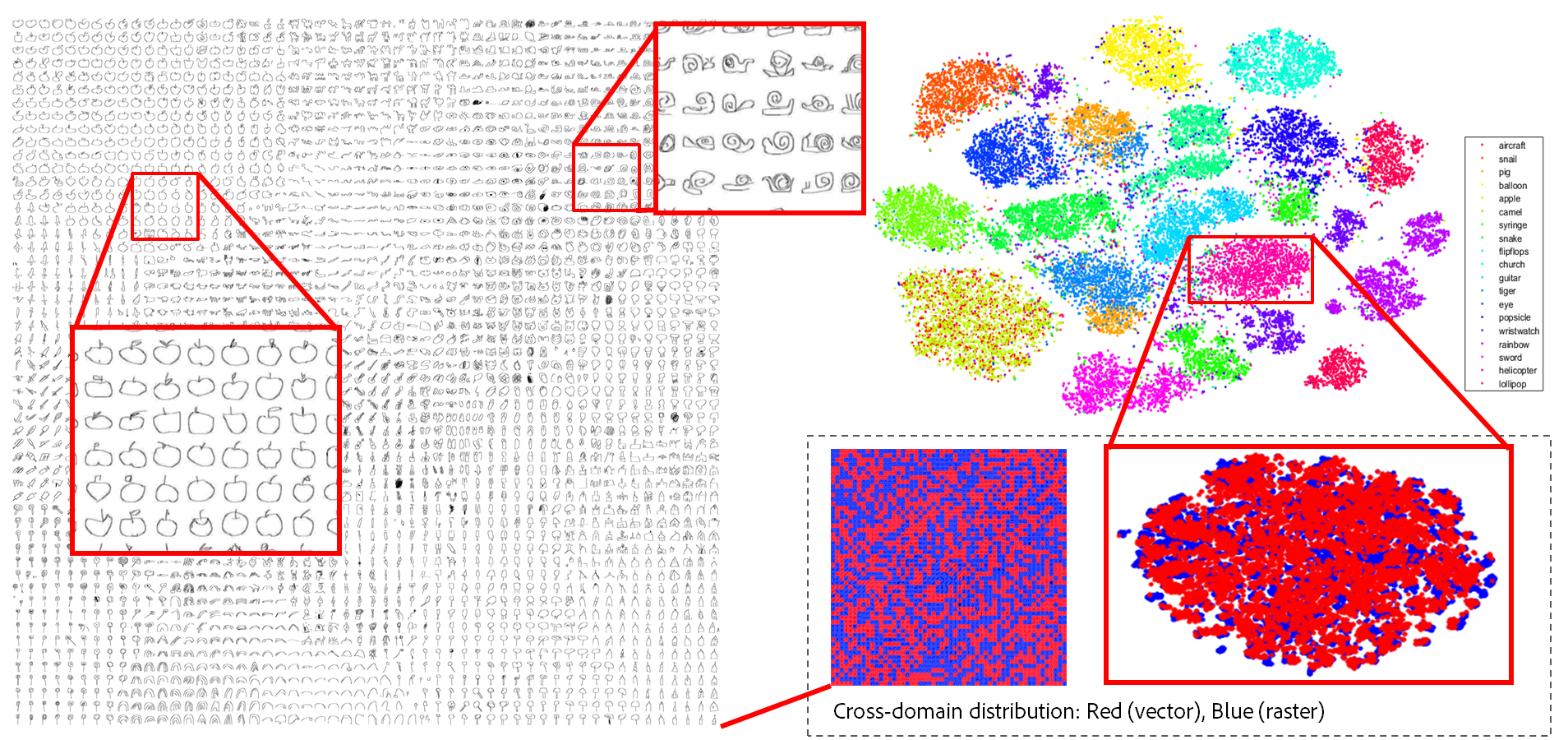}~~~
    (b)~~
    \includegraphics[height=3.75cm]{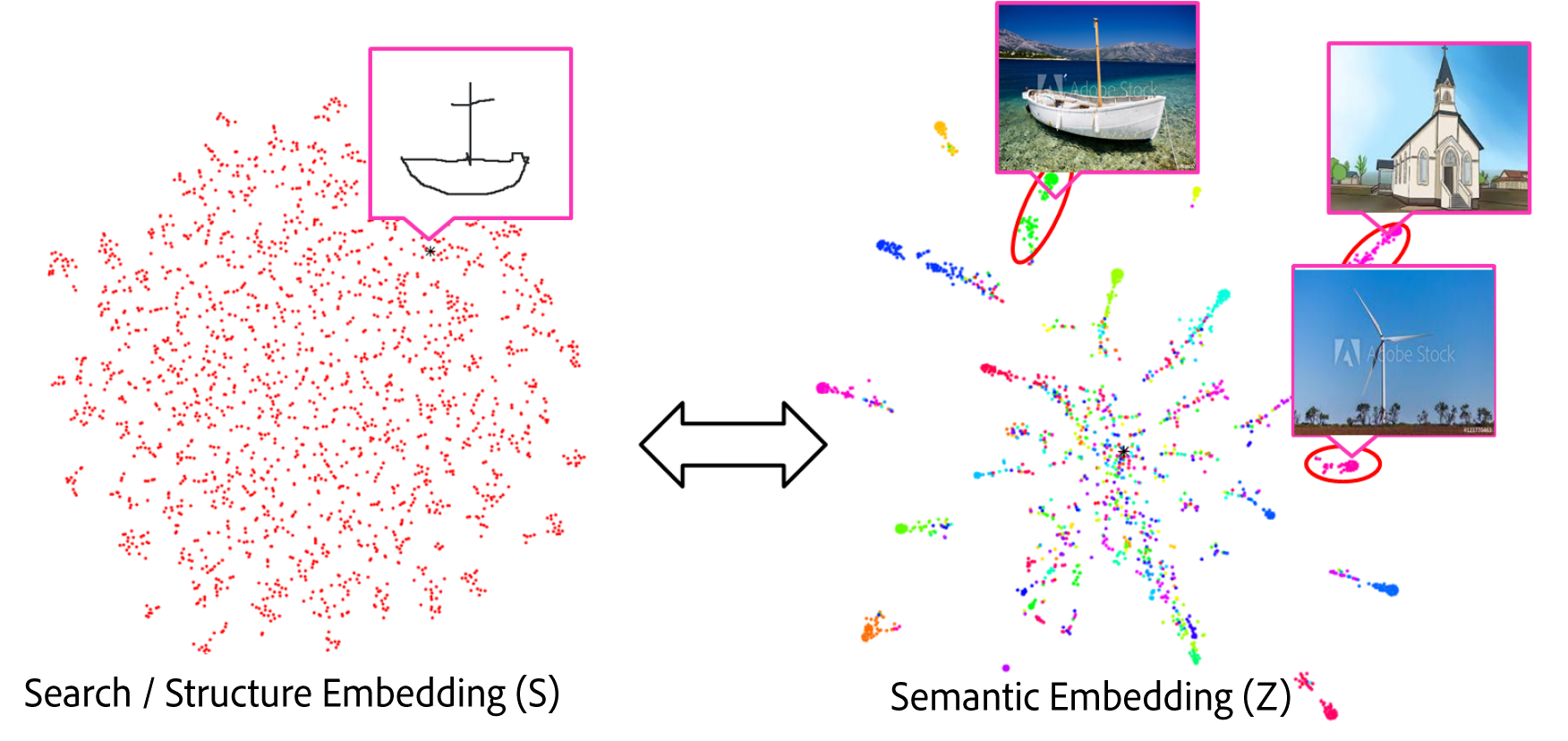}
    \caption{(a) Visualizing the search embedding ($\mathcal{S}$) (for 20/345 random classes sampled from QuickDraw50M); sketches in vector (red) and raster (blue) modalities have been encoded via $S_Q(.)$ and $S_I(.)$ respectively.  The learned representation is discriminative on visual structure but invariant to modality. (b) A k-NN search ($L^2, k=500$) yield search results in $\mathcal{S}$ local to encoded sketch query $S_Q(Q)$; results share similar structure but span diverse semantics \eg a box with a cross atop returns boats, churches, windmills.  Results are clustered in auxiliary (semantic) embedding $\mathcal{Z}$ and presented to user for ranking. }
    \label{fig:vizE}
    \squeezeup
\end{figure*}

The end-to-end triplet network (Fig.~\ref{fig:training}) is trained using sketches only; 3.5M sketches (10K $\times 345$ object classes) sampled from the public Quickdraw50M dataset \cite{qd} (simplified via RDP \cite{rdp}, as in \cite{Eck2018}) and rasterized by rendering pen movements as anti-aliased lines of width 1 pixel on a $256 \times 256$px canvas.  The stroke sequence and the rendering of that sequence are fed through the anchor (a) and positive (p) branches, and a randomly selected rasterized sketch of differing object class to negative branch (n).  Weights are optimized via ADAM using triplet loss computed over activations available from the final shared fc layer (the search embedding $\mathcal{S}$):
\begin{eqnarray}
L_{\mathrm{train}(a,p,n)} &=& [ m + \left||S_Q(a) - F_R(R_S(p))\right||_2^2 -\nonumber \\ 
&& \left||S_Q(a)-F_R(R_S(n))\right||_2^2 ]_+ 
\label{eq:stdtriploss}
\vspace{-10pt}
\end{eqnarray}
  where $m=0.2$ is a margin promoting convergence, and $[x]_+$ indicates the non-negative part of $x$.  Training yields weights for the fully connected (fc) layers -- recall these are partially shared across the vector (a) and raster (p/n) branches, yielding $F_V(.)$ and $F_R(.)$.  The end-to-end
  functions mapping a sketched query $Q$ to our common search embedding $\mathcal{S}$ is:
\begin{equation}
S_Q(Q) = F_V(V_E(Q)).
\label{eq:fc}
\end{equation}
Fig.~\ref{fig:vizE}a shows the resulting embedding; raster and vector content representing similar visual structures mix within $\mathcal{S}$ but distinct visual structures form discriminative clusters.

\subsection{Search Implementation}

Once trained, $S_Q(.)$ forms the RNN path within our search framework (Fig.~\ref{fig:overview}, green) for encoding a vector sketch query $Q$.  The CNN path $S_I(.)$  (Fig.~\ref{fig:overview}, blue) used to index images for search, adopts the {\em image} branch of \cite{Bui2018} (subsec.~\ref{sec:strucnet}):
\begin{equation}
\label{eq:enc_r}
S_I(I) = F_R(R_I(I)).
\end{equation}
Note substitution of the sketch branch $R_S(.)$ that was used during training (eq.\ref{eq:stdtriploss}) for the image branch $R_I(.)$.  Both functions map to the same intermediate embedding $\mathcal{R}$, however we index images rather than sketches for SBIR. 

\subsection{Disambiguating Search Intent}
\label{sec:intent}

Given a search query $Q$, a k-NN lookup within $\mathcal{S}$ is performed to identify a set of results $J=[I_1,...,I_k]$ where $J \subseteq I$ minimising $||S_Q(Q) - S_I(I_i)||_2$; in practice, $||.||_2$ is approximated via product quantization (PQ) \cite{Jegou2010} for scalability and up to $k=500$ results are returned.  The results are clustered into candidate search intents, and presented to the user for feedback.  Clustering is performed within an auxiliary embedding ($\mathcal{Z}$) available from final layer activations of a ResNet50/ImageNet pre-trained CNN. We write this function $Z(I_i)$, pre-computed $\forall I_i \in I$  during indexing.

\subsubsection{Clustering}
Images local to $S_Q(Q)$ within the search embedding $\mathcal{S}$ may be semantically diverse; a single visual structure \eg a cross atop a box, may return churches, boats, windmills, etc.  However these results will form distinct clusters within $\mathcal{Z}$ (Fig.~\ref{fig:vizE}b).  We apply affinity propagation \cite{afprop} to identify the dominant $m=3$ clusters $C=[c_1,...,c_m]$ in $\mathcal{Z}$.  The algorithm constructs an affinity graph for all image pairs in $(I_a,I_b) \in J \times J$ scoring these:
\begin{equation}
d(I_a,I_b)=||Z(I_a) -Z(I_b)||_2.
\end{equation}
Clustering is a greedy process that iteratively constructs $C$, selecting a best cluster $c_i={I_1,...,I_k}$ from the graph, minimizing $\rho(c_i)$:
\begin{equation}
\rho(c_i) = \sum_{(I_a,I_b) \in c_i \times c_i}  d(I_a,I_b) + W(c_i,C).
\end{equation}
Where $W(C)$ is a penalty term that encourages semantic diversity by discouraging selection of $c_i$ containing images similar to those already in $C$:
\begin{equation}
W(c_i,C) \propto -log \left(\sum_{(I_a,I_b) \in c_i \times \chi(C)} d(I_a,I_b)\right)
\end{equation}
where $\chi(c_i)$ represents the set of images already present within clusters in set $C$.

\subsubsection{Identifying  Search Targets}

For each cluster $i=[1,m]$ we identify a representative image $I_i^*$ closest to the visual structure of the query:
\begin{eqnarray}
I_i^* = \argmin_{I_j} ||S_Q(Q)-S_I(I_j)||_2;~~\forall I_j \in C_i.
\end{eqnarray}

Leveraging the Quickdraw50M dataset (QD-3.5M) of sketches ($H$), we identify the closest sketch $Q_i^*$ to each representative image:
\begin{eqnarray}
Q_i^* = \min_{q \in H} ||S_Q(q)-S_I(I_i^*)||_2.
\end{eqnarray}

The set of these sketches $T=\{T_1,...,T_m\}$, where $T_i=Q_i^*$, represent the set of search targets and the basis for perturbing the user's query ($Q$) to suggest a new sketch $Q'$ that guides subsequent iterations of search.

\subsection{Sketch Perturbations for User Guidance}
\label{sec:perturb}

The search targets $T$ are presented to the user, alongside sliders that enable the relevance of each  to be interactively expressed as a set of weights $\Omega=\{\omega_1,...,\omega_m\}$.   We seek a new sketch query $Q'$ that updates the original query $Q$ to resemble the visual structure of those targets, in proportion to these user supplied weights.  

For brevity we introduce the following notation. $Q_i^{V*}=V_E(Q_i^*)$ describes each search target within the RNN embedding $\mathcal{V}$.  Similarly  $Q_i^{S*}=S_Q(Q_i^*)$ describes each target within the search embedding $\mathcal{S}$.  We similarly use $Q^V=V_E(Q)$ and $Q^S=S_Q(Q)$ to denote the user's sketch $Q$ in $\mathcal{V}$ and $\mathcal{S}$ respectively.   To perturb the sketch we seek $Q'$ (similarly written $Q^{V'}$ and $Q^{S'}$ within those embeddings).

The availability of the v.a.e. decoder (subsec.~\ref{sec:sketchrnn}) enables generation of new sketches (sequences of strokes) conditioned on any point within $\mathcal{V}$.  Our approach is to seek $Q^{V'}$ such that $Q'=V_D(Q^{V'})$ may be generated.  The task of updating $Q \mapsto Q'$ is therefore one of obtaining $Q^{V'}$ via interpolation between $Q^V$ and targets $Q_i^{V*}$, as a function of user supplied weights and targets.
\begin{equation}
Q^{V'}=f(Q^{V};\Omega,T).
\label{eq:gen}
\end{equation}
We describe two strategies (instances of $f$) for computing $Q^{V'}$ from query $Q^V$ (evaluating these in subsec.~\ref{sec:eval_interp}).

\begin{figure}[t!]
    \centering
    \includegraphics[width=1.0\linewidth]{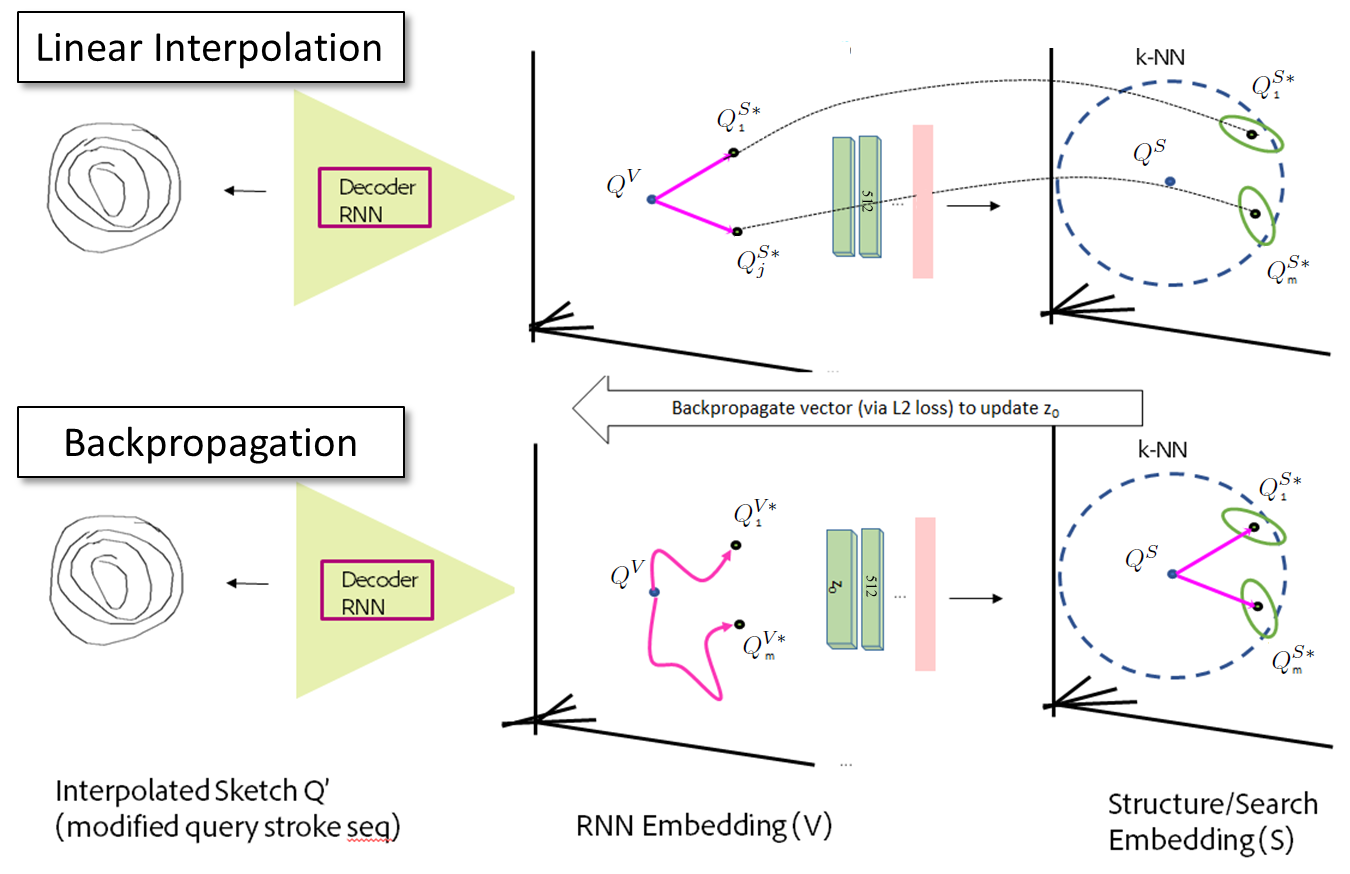}
    \caption{Generating $Q'$:  Linear (top) vs. non-linear (bottom) interpolation in RNN space ($\mathcal{V}$); the latter due to backpropagation of loss eq.~\ref{eq:bploss} that updates $Q\mapsto Q'$ s.t. $Q^{S'}$ tends toward the search targets identified local to $Q^S$ by the user. See Fig.~\ref{fig:interp} for examples.}
    \label{fig:blend}
    \squeezeup
\end{figure}

\subsubsection{Linear Interpolation}
A na\"ive solution is to linearly interpolate within the RNN embedding $\mathcal{V}$, \ie:
\begin{equation}
f_\mathrm{linear}(Q^V;\Omega,T)= Q^V + \sum_{i=1}^m \omega_j (Q_i^{V*}-Q^V)
\end{equation}

yielding $Q^{V'}$ via eq.~\ref{eq:gen}, from which the sketch suggestion $Q'$ is generated via RNN decoder $Q'=V_D(Q^{V'})$. However, although $Q^S$ and $Q_i^{S*}$ are local by construction, it is unlikely that $Q^V$  and $Q_i^{V*}$ will be local; nor is the manifold of plausible sketches within $\mathcal{V}$ linear.    This can lead to generation of implausible sketches (c.f. Sec.~\ref{sec:eval_interp}, Fig.~\ref{fig:interp}).  

\subsubsection{Back-propagation}

We therefore perform a non-linear interpolation in $\mathcal{V}$, minimizing an objective that updates $Q^{V} \mapsto Q^{V'}$ closer to search target(s) $Q_i^{V*}$ via backprop through the fc layers $F_V(.)$  (eq.\ref{eq:fc}) to reduce the distance between $Q^S$ and $Q_i^{S*}$.  
\begin{eqnarray}
\label{eq:dst_q}
D(Q^{V'})&=& \frac{1}{m}\sum_{j=1}^{m} \omega_j||Q^{S'}- Q_j^{S*}||_2^2.
\end{eqnarray}
This is analogous to FGSM adversarial perturbation (AP) of images in object recognition \cite{Goodfellow2015}, where the input the network is modified via backprop to influence its mapping to a classification embedding. In our context, we define a loss based on this distance in $\mathcal{S}$  regularized by the constraint that the original and updated sketch should be nearby in $\mathcal{V}$:
  \begin{eqnarray}
\label{eq:bploss}
L_{\mathrm{AP}}(Q')&=&D(Q^{V'}) + \alpha||Q^{V'} - Q^{V}||_2.
\end{eqnarray}
Weight $\alpha=0.1$ was emperically set. An optimal $Q^{V'}$ is sought by backpropagation through the fc layer $F_V(.)$:
\begin{equation}
f_\mathrm{AP}(Q';\Omega,T)= \argmin_{q'}L_{\mathrm{AP}}(V_E(q')).
\end{equation}
Equipped with sliders to control relevance weights $\Omega$ on each of the targets, the user in effect performs a linear interpolation (between $S_Q(Q)$ and $T$) within $\mathcal{S}$ that causes a non-linear extrapolation from $Q^{V}$ to output point $Q^{'V}$, and ultimately the sketch suggestion via  RNN decoder $Q'=F_D(Q^{'V})$.  Fig.~\ref{fig:blend} contrasts the linear and non-linear (backprop) approaches; visual examples in Fig.~\ref{fig:interp}.

\section{Experiments and Discussion}
\label{sec:results}

We evaluate the performance of the LiveSketch using the QuickDraw50M dataset \cite{qd} and a corpus of 67M stock photo and artwork images (Stock67M) from Adobe Stock\footnote{Downloaded from \url{https://stock.adobe.com} in late 2016}.

{\bf QuickDraw50M} is a dataset of 50M hand-drawn sketches crowdsourced via a gamified classification exercise  (Quick, Draw!) \cite{qd}. Quickdraw50M is well suited to our work due to its class diversity (345 object classes), vector graphics (stroke sequence) format, and the casual/fast, throwaway act of the sketching encouraged in the exercise that reflects typical SBIR user behaviour \cite{Collomosse2008} (vs. smaller, less category-diverse datasets such as TUBerlin/Sketchy that contain higher fidelity sketches drawn  with reference to a target photograph \cite{Hays2016,Eitz2012}). We sample 3.5M sketches randomly with even class distribution from the Quickdraw50M training partition to create training set ({\bf QD-3.5M}; detail in  subsec~\ref{sec:learnfc}).  For sketch retrieval and interpolation experiments we sample 500 sketches per class (173K total) at random from the Quickdraw50M test partition to create an evaluation set {\bf QD-173K}).   A query set of sketches ({\bf QD-345}) is sampled from QD-173K, one sketch per object class, to serve as queries for our non-interactive experiments.

{\bf Stock67M} is a diverse, unannotated corpus of images used to evaluate large-scale SBIR retrieval performance.  The dataset was created by scraping harvesting every public, unwatermarked image thumbnails from the Adobe Stock website in late 2016 yielding approximately 67M images at QVGA resolution.

\begin{figure}[t!]
    \centering
    \includegraphics[width=0.95\linewidth,height=4.5cm]{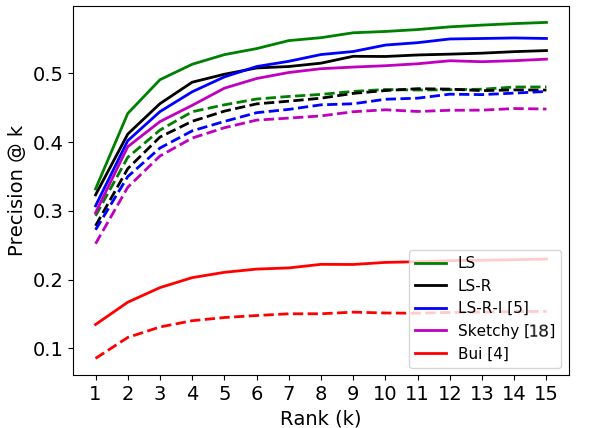}
    \includegraphics[width=1.0\linewidth,height=4.5cm]{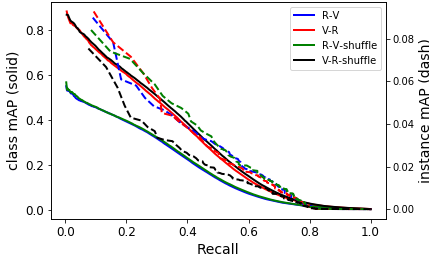}
    \caption{Performance of joint search embedding for retrieval. Top: Sketch2Image (S-I) Precision@k curve for SBIR -- see Tbl.~\ref{tbl:retrieval} for mAP\% and key to ablation notation. Bottom: Sketch2Sketch (S-S) cross-modal matching.  Class-level (solid) and instance level (dash) mAP-recall curves for Vector-Raster (V-R), Raster-Vector (R-V), and stroke shuffling experiments (-shuffle). }
    \label{fig:curves}
    \squeezeup
\end{figure}

\subsection{Evaluating cross-modal search}
\label{sec:eval_embed}

We evaluate the performance of our cross-modal embedding ($\mathcal{S}$) for sketch based retrieval of sketches and images.

\begin{table}[t]
\centering
\begin{tabular}{l|c|cc|}
\toprule 
    ~ & Method & Class-level & Instance-level\\
    \midrule
    \multirow{4}{*}{S-I}& LS {\em (Ours)} & \bf{38.40}  & \bf{30.81} \\
& LS-R & 35.26  & 29.48 \\
& LS-R-I \cite{Bui2018}& 35.15  & 27.48 \\
& Sketchy \cite{Hays2016}& 33.21  & 27.06 \\
& Bui \etal \cite{Bui2017}& 12.59  & 8.76 \\
\midrule
    \multirow{4}{*}{S-S}& V-R & 34.88  & {\bf 18.80} \\
& R-V & 29.31  & 18.29 \\
& V-R-shuffle  & {\bf 35.94}  & 15.71 \\
& R-V-shuffle  & 29.61  & 18.57 \\
\bottomrule
\end{tabular}
\caption{Accuracy for sketch based recall of sketches (S-S) and images (S-I); evaluated using class and instance level mAP (\%) over 345 vector queries (QD-345). Top: S-I ablations; raster query (LS-R) and raster intermediate embedding (LS-R-I) \cite{Bui2018}.  Bot. S-S retrieval across query modalities; raster querying vector (R-V) and vector querying raster (V-R); also variants stroke order (-shuffle).}
\label{tbl:retrieval}
\squeezeup
\end{table}

\noindent \textbf{Sketch2Sketch (S-S) Matching.} We evaluate the ability of our embedding trained in subsec.~\ref{sec:learnfc} to discriminate between sketched visual structure, invariant to input modality (vector vs. raster). We train our model on QD-3.5M and retrieve sketches from the QD-173K corpus, using QD-345 as queries.   Both vector queries retrieving raster content (V-R) and vice versa (R-V) are explored, using category (class-level) and fine-grain (instance-level) metrics.  For the former we consider a retrieved record a match if it matches the sketched object class.  For the latter, the exact same sketch must be returned. To run raster variants, a rasterized version of QD-173K is produced by rendering strokes to a $256 \times 256$ pixel canvas (see method of subsec.~\ref{sec:learnfc}).  Sketches from QD-173K are encoded to the search embedding $\mathcal{S}$  from their vector and raster form respectively via functions $S_Q(.)$ and $F_R(R_S(.))$.  Fig.~\ref{fig:vizE} visualizes the sketches within the search embedding; similar structures cluster together whilst the vector/raster modalities mix. Tbl.~\ref{tbl:retrieval} (bot.) and Fig.~\ref{fig:curves} (bot.) characterize performance; vector queries ($\sim 35\%$ mAP) outperform raster ($\sim 29\%$ mAP) by a $\sim 5\%$ margin.  To explore this gain further, we shuffled the ordering of the vector strokes retraining the model from scratch. We were surprised to see comparable performance at class-level, suggesting this gain is due to the spatial continuity inherent to the stroke representation, rather than temporal information.  The fractional increase may be due to the shuffling acting as data augmentation enhancing invariance to stroke order. However when at instance level, ordering appears more significant ($\sim 3\%$ gain; V-R vs. V-R-shuffle).

\begin{figure}[t!]
    \centering
    \includegraphics[width=1.0\linewidth,height=4.5cm]{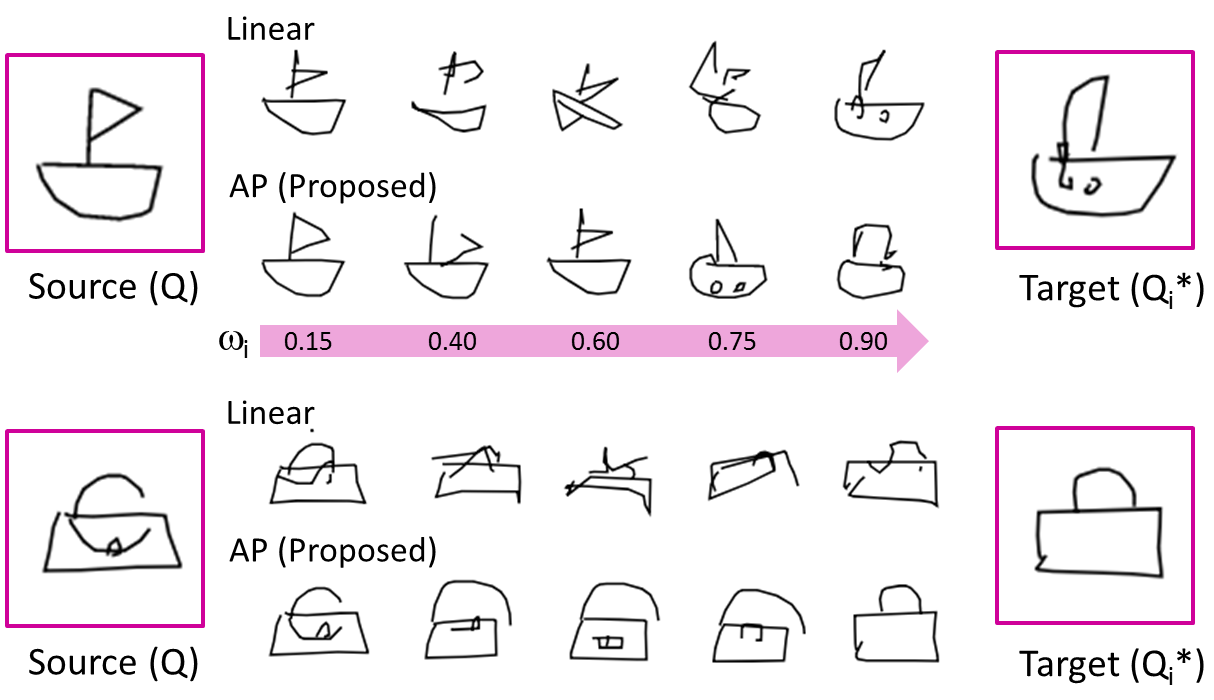}
    \caption{Comparison of sketch interpolation via our backprop ($f_{AP}$) approach and linear interpolation ($f_{linear}$) within $\mathcal{V}$, for fine-grain variations of a boat and a bag (c.f. Tbl.~\ref{tbl:interp}).}
    \label{fig:interp}
    
    \squeezeup
\end{figure}

\begin{table}[t!]
{
\centering
    \includegraphics[width=1.0\linewidth]{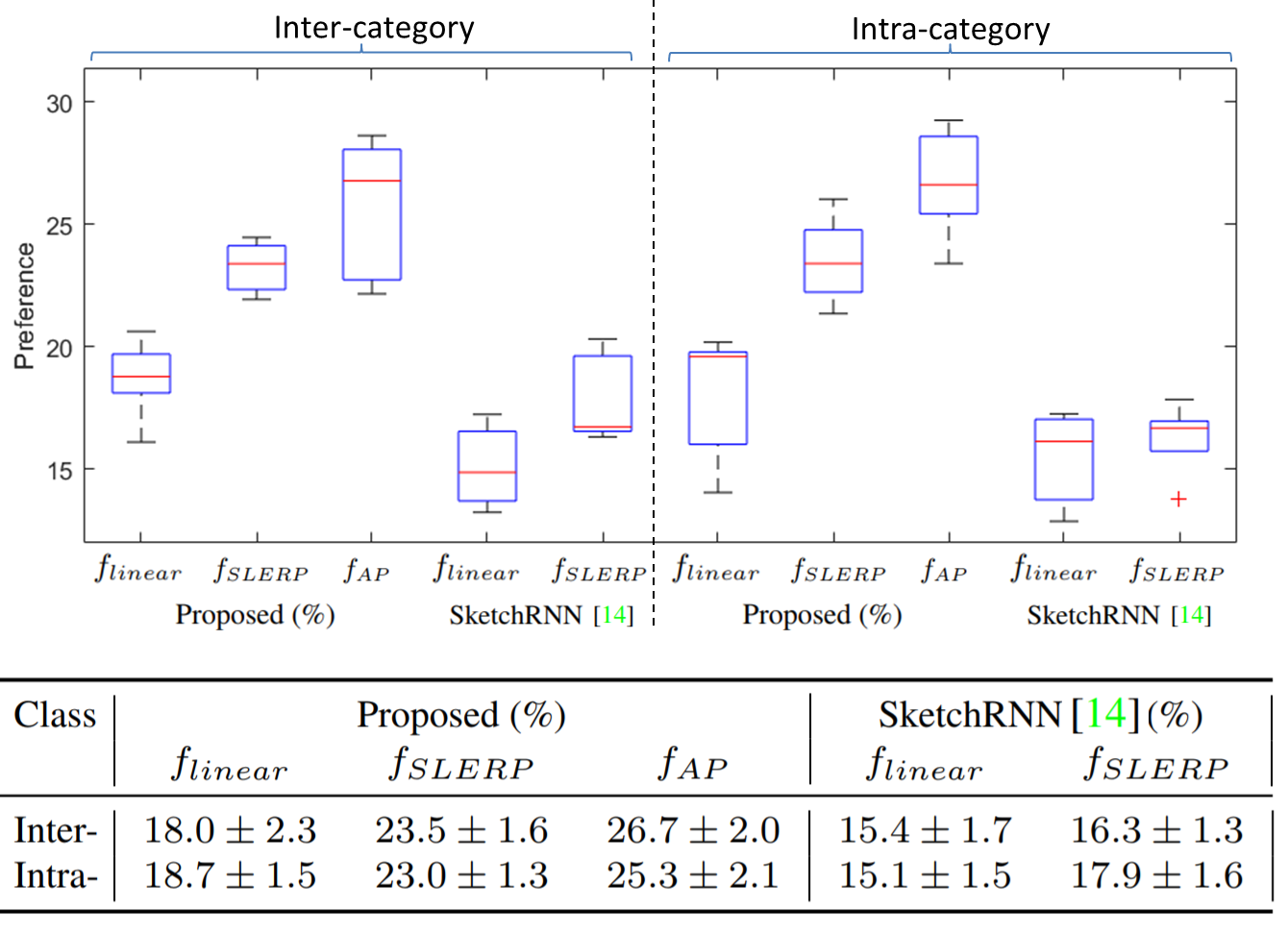}
%
}
\vspace{-1em}
\caption{Perturbation method user study (MTurk) comparing the proposed query perturbation scheme inspired by adversarial perturbations ($f_{AP}$) vs. linear interpolation variants ($f_{linear},f_{SLERP}$).}
\label{tbl:interp}
\squeezeup
\end{table}

\begin{table*}[t!]
{
\centering
~~~~~~~\begin{tabular}{|l|ccc|ccc|}
\toprule 
Method & \multicolumn{3}{c}{Ablations: $seconds (missed)$} & \multicolumn{3}{c}{Baselines $seconds (missed)$} \\
    & LS (Ours) & LS-NI & LS-NI-R & LS-NI-R-I \cite{Bui2018} & Sketchy \cite{Hays2016} & Bui \etal \cite{Bui2017}\\
    \midrule
Class-level T-T & 24.90 (1.33)  & 38.33 (1.33) & 31.74 (0.33)  & \bf{19.12 (0.00)} &  46.20 (1.00)  & 40.13 (1.33)  \\     
Instance-level T-T & \bf{30.74 (2.00)}  & 45.43 (1.67)  & 66.46 (3.67) & 95.27 (3.67)  & 80.28 (2.67)  & 75.02 (1.33)\\     \midrule
Mean Avg. T-T & \bf{27.67 (3.33)}  & 41.72 (3.00)  & 45.92 (4.00) & 42.69 (3.67)  & 60.90 (3.67)  & 54.88 (2.67)\\   \bottomrule 
\end{tabular}
}
\caption{Time-to-task user study. Average time to retrieve 20 class- and instance-level search targets (18 participants, 3 per method).  Comparing LiveSketch (LS) interactive method with ablations (-NI) non-interactive/one-shot; (-R) raster substitutes vector query; (-I) intermediate structure embedding, and with the three baselines \cite{Bui2018,Hays2016,Bui2017}. Times in seconds; parentheses  total the averaged missed queries. }
\label{tbl:userstudy}
\squeezeup
\end{table*}
\noindent \textbf{Sketch2Image (S-I) Matching.} We evaluate the performance of the search embedding for SBIR over Stock67M, using all QD-345 sketch queries (without user interaction in this experiment).  Annotation of 67M images for every query is impractical; we instead crowd-source per-query annotation via Mechanical Turk (MTurk) for the top-$k$ ($k$=15) results and compute both mAP\% and precision@$k$ curve averaged across all 345 queries for each experiment. The annotation is crowd-source with 5 repetitions. Results are summarized in  Tbl.~\ref{tbl:retrieval} (S-I) and Fig.~\ref{fig:curves} (top).  We perform two ablations to our proposed LiveSketch (LS) system: 1) querying with rasterized versions of the QD-345 queries (-R) using the proposed embedding $\mathcal{S}$; 2) querying with rasterized queries in the intermediate embedding $\mathcal{R}$ (-R-I) which degenerates to \cite{Bui2018}; we also baseline against two further recent SBIR techniques: the unshared triplet GoogleNet-V1 architecture proposed by Sangkloy \etal \cite{Hays2016}, and the triplet edgemap approach of Bui \etal \cite{Bui2017}.  We compute class- and instance- level precision for all queries resulting in $345 \times 15 \times 5 = \sim26\mathrm{K}$ MTurk annotations. Our embedding (LS) outperforms all ablations and baselines, with vector query alone contributing significant margin over raster.  The addition of fc layers to create cross-modal embedding (-R) slightly improves (importantly, does not degrade) the  intermediate raster embedding $\mathcal{R}$ available via \cite{Bui2018}.  The method significantly outperforms recent triplet SBIR approaches \cite{Hays2016,Bui2017}.  Note that the S-I and S-S figures are non-comparable; they search different datasets.

\subsection{Evaluating Search Suggestions}
\label{sec:eval_interp}

MTurk was used to evaluate the relative performance of sketch interpolation techniques used to form query suggestions ($Q'$). We benchmark linear ($f_{linear}$) and spherical linear (SLERP, \cite{Eck2018} $f_{SLERP}$) interpolation\cite{Eck2018} in our  RNN embedding $\mathcal{V}$ with the proposed approach $f_{AP}$ inspired by adversarial perturbations, in which the sketch is perturbed via non-linear interpolation in $\mathcal{V}$ due to backpropagation.  We also compare to linear and SLERP embedding within the embedding of the original SketchRNN network of Eck \etal \cite{Eck2018} trained using the same data (QD-3.5M).

MTurkers were presented with a pair of sketches $Q$ and $Q'$ sampled from QD-173K and shown a sequence of 10 sketches produced by each of the 5 interpolation methods.  MTurkers were asked to indicate which interpolation looked most natural / human drawn.  Each experiment run  sampled 300 intra- and 300 inter-category pairs $(Q,Q')$ picked at random from QD-173K. The experiment was repeated 5 times yielding 3k annotations from 25 unique MTurkers.

Tbl.~\ref{tbl:interp} summarizes user study results; an un-paired t-test \cite{ttest} was run to determine significance ($p$).  Backpropagation ($f_{AP}$, proposed) outperformed direct linear interpolation ($f_{linear}$) in $V$ for inter- (18.0\% vs. 26.7\%, $p<0.002$) and intra-category (18.7\% vs. 25.3, $p<0.030$) cases (see Fig.~\ref{fig:interp} for visual examples).  Statistically significant results were  obtained for $f_{SLERP}$ at $p<0.03$.  In both cases preference was stronger for inter-category interpolation, likely due to non-local nature of ($Q,Q'$) causing linear interpolation to deviate from the manifold of plausible sketches  (enforced by $f_{AP}$). Even linear interpolation in $\mathcal{V}$ enabled more natural interpolations in both inter- and intra-category cases vs. original SketchRNN \cite{Eck2018}; but this was significant only for the former.


\squeezeupSmall
\subsection{Evaluating Iterative Retrieval}
\label{sec:eval_system}

We evaluate the efficacy of LiveSketch via a time-to-task experiment in which  18 participants were timed searching for 20 targets using 6 methods. We perform 3 ablations to our method (LS): 1) non-interactive (-NI), users are not offered sketch suggestions; 2) sketches are rasterized (-R) rather than processed as vector queries; 3) as -R but searching within intermediate embedding $\mathcal{R}$  which degenerates to \cite{Bui2018} (-R-I). We  also baseline against \cite{Hays2016,Bui2017}. 
 
Fig.~\ref{fig:teaser} provides a representative query, suggestions and clustered results sampled from the study.  Tbl.~\ref{tbl:userstudy} summarizes the results, partitioning across class- and instance- level queries (10 each). Class-level (category) queries prompted the user to search for a specific object (`{\em cruise ship sailing on the ocean}').  Instance-level (fine-grain) queries prompted the user to search for a specific object with specific pose or visual attributes (`{\em church with three spires}',`{\em side view of a shark swimming left}').   Timing began on the first stroke drawn and ended when the user was satisfied that the target had been found (self-assessed). If a user took longer than three minutes to find a target, then the search time was capped and noted as a miss (bracketed in Tbl.~\ref{tbl:userstudy}).

Significant decrease in time-to-task ($\sim 15$s) was observed with the  interactive method (LS) over non-interactive variants using vector (LS-NI, LS-NI-R, LS-NI-R) although query modality had negligible effect on mean time to task for non-interactive cases.  Baselines performed $\sim 10-20$s slower onexplainable via lower retrieval performance in subsec.~\ref{sec:eval_embed}.  In all cases, fine-grain queries took longer to identify with greater instances of missed searches -- however the margin over class-level searches was only $\sim 6$s vs. interactive.  Whilst category level search time was not enhanced by the proposed method, time taken to produce successful fine-grain sketch queries was significantly reduced by up to $15$s over non-interactive ablations and by a factor of 3 over baselines.  All 6 methods used a PQ \cite{Jegou2010} index and took $30-40ms$ to run each query over the Stock67M corpus.

\section{Conclusion}
\label{sec:conclude}
We presented an novel architecture that, for the first time, enables search of large image collections using sketched queries expressed as a variable length sequence of strokes (\ie in 'vector' form).  The vector modality enables a seocnd contribution; live perturbation of the sketched query strokes to guide the user's search.  The search is guided towards likely search intents --- interactively specified by user weights attributed to clusters of results returned at each search iteration.  Perturbations were generated via backpropagation through the feature encoder network, inspired by adversarial perturbations (FGSM \cite{Goodfellow2015}) typically used to attack object classification systems,  and applied for the first time here in the context of relevance feedback for visual search. We showed that our interactive system significantly reduces search time over a large (67M) image corpus, particularly for instance-level (fine-grain) SBIR, and that our search embedding (unifying vector/RNN and image/CNN modalities) performs competitively against three baselines \cite{Bui2018,Hays2016,Bui2017}.  Future work could focus on improvement of the RNN embedding which can still produce implausible sketches for very detailed (high stroke count) drawings. 

{\small
\bibliographystyle{ieee_fullname}
\bibliography{livesketch}
}

\end{document}